\documentclass[letterpaper, 10 pt, conference]{ieeeconf}  % Comment this line out if 

\usepackage{pifont}
\usepackage{graphicx}
\usepackage{subfigure}

\usepackage{amsmath}
\usepackage{mathtools}
\usepackage{algorithm}
\usepackage{amssymb}
\usepackage{xcolor}
\usepackage{algpseudocode}

\IEEEoverridecommandlockouts                     

\title{\LARGE \bf
Swarm Formation Morphing for Congestion Aware Collision Avoidance*}

\author{Jawad N. Yasin$^{1}$, Mohammad-Hashem Haghbayan$^{1}$, Muhammad Mehboob Yasin$^{2}$ and Juha Plosila$^{1}$% <-this % stops a space
\thanks{*This work has been supported in part by the Academy of Finland-funded research project 314048 and Nokia Foundation.}% <-this % stops a space
\thanks{$^{1}$Jawad N. Yasin, Mohammad-Hashem Haghbayan, and Juha Plosila are with the the  Autonomous  Systems  Laboratory, Department of Future Technologies, University of Turku, 20500, Finland
        {\tt\small \{janaya, mohhag, juplos\}@utu.fi}}%
\thanks{$^{2}$Muhammad Mehboob Yasin is with the Department of Computer Networks, College of Computer Sciences \& Information Technology, King Faisal University, Hofuf, Saudi Arabia
        {\tt\small mmyasin@kfu.edu.sa}}%
}

\begin{document}

\maketitle
\thispagestyle{empty}
\pagestyle{empty}

%%%%%%%%%%%%%%%%%%%%%%%%%%%%%%%%%%%%%%%%%%%%%%%%%%%%%%%%%%%%%%%%%%%%%%%%%%%%%%%%
\begin{abstract}

The focus of this work is to present a novel methodology for optimal distribution of a swarm formation on either side of an obstacle, when evading the obstacle, to avoid overpopulation on the sides to reduce the agents' waiting delays, resulting in a reduced overall mission time and lower energy consumption. To handle this, the problem is divided into two main parts: 1) the disturbance phase: how to morph the formation optimally to avoid the obstacle in the least possible time in the situation at hand, and 2) the convergence phase: how to optimally resume the intended formation shape once the threat of potential collision has been eliminated. For the first problem, we develop a methodology which tests different formation morphing combinations and finds the optimal one, by utilizing trajectory, velocity, and coordinate information, to bypass the obstacle. For the second problem, we utilize a thin-plate splines (TPS) inspired temperature function minimization method to bring the agents back from the distorted formation into the desired formation in an optimal manner, after collision avoidance has been successfully performed. Experimental results show that, in the considered test scenario, the traditional method based on the shortest path results in 14.7\% higher energy consumption as compared to our proposed approach.

\end{abstract}

%%%%%%%%%%%%%%%%%%%%%%%%%%%%%%%%%%%%%%%%%%%%%%%%%%%%%%%%%%%%%%%%%%%%%%%%%%%%%%%%
\section{Introduction}

\begin{figure}[!ht]
    \centering
    \includegraphics[width=0.35\textwidth]{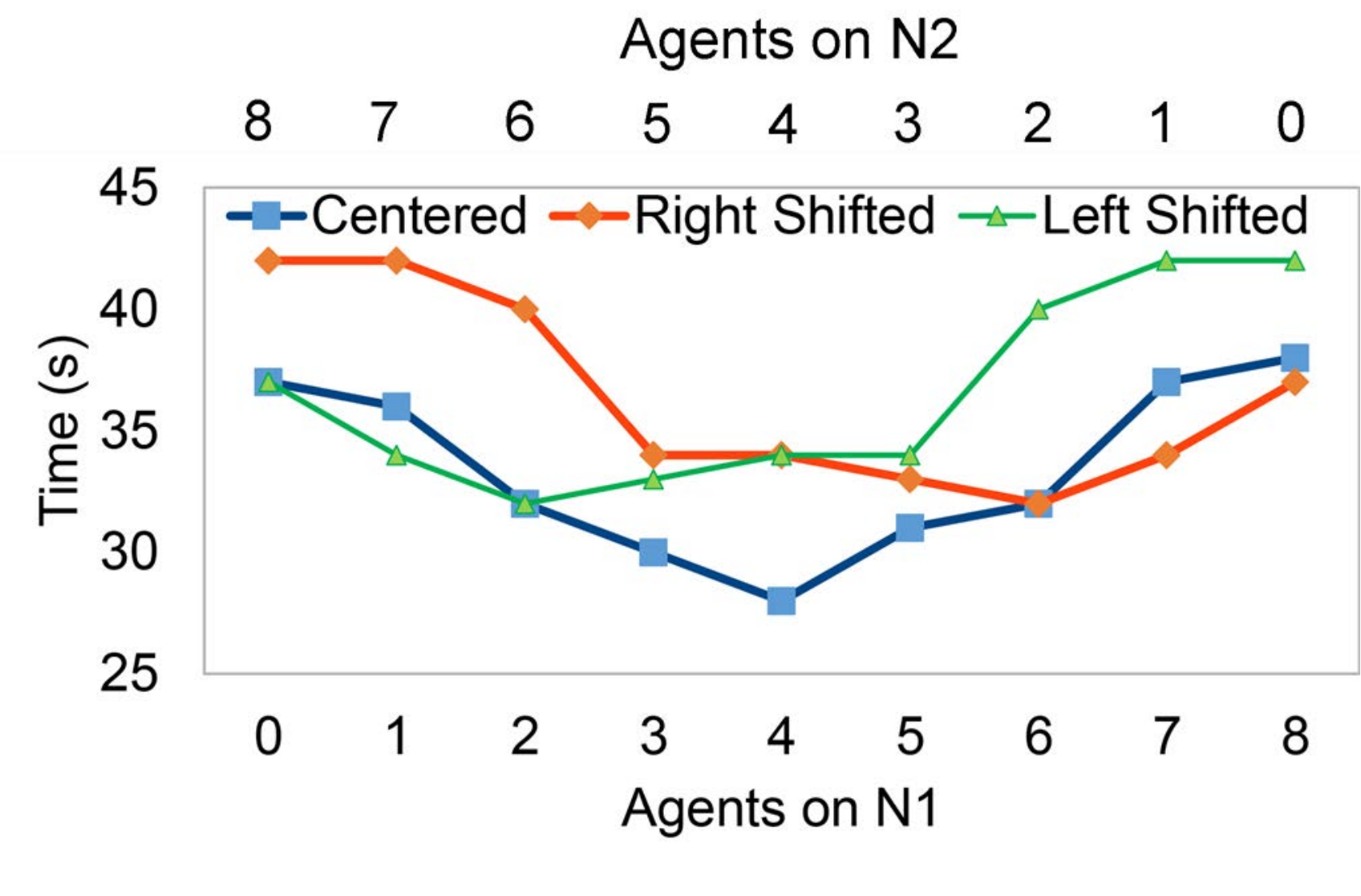}
    \caption{Obstacle's location w.r.t. the swarm: Centered = Obstacle's center is in the same as the swarm's center, Left Shifted = Obstacle is shifted towards left side from the center of the swarm, Right Shifted = Obstacle is shifted towards right side from the center of the swarm\label{fig:responsegraph}}
\end{figure}

The study of the behaviour of a system comprising a large number of autonomous agents that interact amongst themselves as well as the environment is generally classified as swarm robotics \cite{Hamann2018, dorigo2004swarm}. Study of swarms of UAVs (drones) has seen rising interest from research community due to their integration in diverse application fields, such as transportation \cite{7989678}, atmospheric research \cite{8682048}, surveillance \cite{1678135}, entertainment \cite{6385551}, and mapping in GPS-denied environments \cite{8764393}, due to their ability to work in collaborative and cooperative manner \cite{9197184}. Navigation of a swarm of agents introduces several research challenges. Among these, the two most significant ones are formation maintenance and collision avoidance \cite{jdj2}. Collision avoidance systems are responsible for guiding an autonomous agent in order to safely and reliably avoid potential collisions with other agents in the swarm as well as with other objects in the environment \cite{jdj1}. Reducing energy consumption to increase mission life is another important research area in swarm robotics, focusing on a diverse set of topics, such as efficient decision making \cite{7850263}, minimization of travelling distance \cite{Majd_2020}, energy efficient communication for swarm robot coordination \cite{8711119}, decreasing the usage of ranging sensors \cite{10.1007/978-3-030-49778-1_28}, and autonomous recharging \cite{tseng2017autonomous}. Current approaches of minimizing the travelling distance while performing collision avoidance maneuvers may adversely impact energy efficiency of the swarm owing to congestion on narrow pathways. This serves as a key motivation for the approach proposed in this paper. As shown in Figure \ref{fig:responsegraph} and elaborated in Section 5, collision avoidance maneuvers focusing on time minimization of the swarm as a whole is a more effective method.
\begin{figure*}[!ht]
\begin{center}
\vspace*{0.2cm}
    \subfigure[\label{fig.eg1}]{\includegraphics[width=0.5\columnwidth]{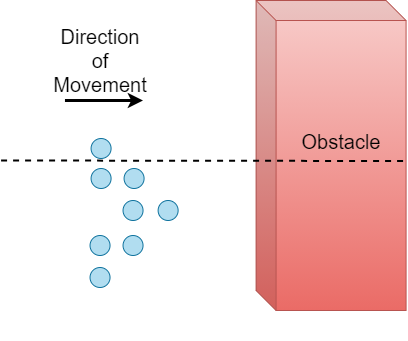}}
    \subfigure[\label{fig.eg2}]{\includegraphics[width=0.5\columnwidth]{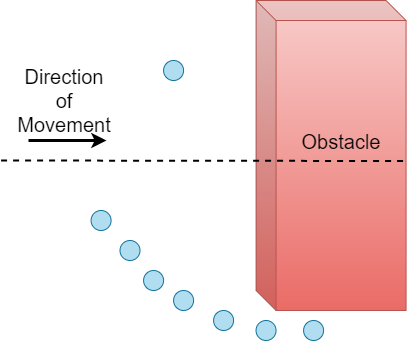}}
    \subfigure[\label{fig.eg3}]{\includegraphics[width=0.5\columnwidth]{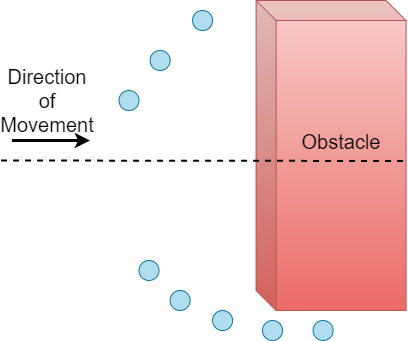}}
    %\subfigure[\label{fig:sys}]{\includegraphics[width=0.5\columnwidth]{overallsys.png}}
   \end{center}\vspace{-0.15cm}
  \caption{Swarm encountering obstacle (a) the initial configuration, (b) shortest path swarm distribution, (c) swarm distribution  utilizing the proposed approach}\vspace{-0.3cm}
  \label{fig:egfig}
\end{figure*}
When a swarm of autonomous drones encounters an obstacle(s), the agents take local decisions to perform collision avoidance maneuvers. Figure \ref{fig:egfig} shows an example scenario of a swarm with eight agents avoiding an obstacle using the two different approaches. The initial configuration is illustrated by agents in "blue" (Figure \ref{fig.eg1}). The cases illustrated are as follows: 1) swarm in distribution while performing collision avoidance using shortest path approach (Figure \ref{fig.eg2}), 2) the distribution of the swarm agents with the proposed approach (Figure \ref{fig.eg3}). The apparent answer to the collision avoidance problem is for each drone to select the nearest end of the obstacle and go round the corner as the optimum route, namely: the shortest path approach \cite{8284558}. As exemplified in the aforementioned figure, the optimal formation disturbance for the swarm may not follow the shortest path rule, for example in Figure \ref{fig.eg2} if each agent moves towards the edge of the obstacle with respect to its own coordinates to follow the shortest path, it will take more time for the swarm to bypass the obstacle since the agents will have to slow down to avoid congestion from neighboring agents. On the other hand, if the agents follow the proposed optimal morphing configuration, illustrated in \ref{fig.eg3}, the swarm distribution is done in a manner to minimize the overall time penalty. In order to avoid the congestion and resultant delays, some of the agents are directed to choose longer the routes in order to minimize the overall time taken by the swarm to pass the obstacle.

\section{Problem formulation}

The main motivation behind the proposed approach comes from the hypothesis that the selection of avoidance route may apparently be the shortest path whereas it may not be the optimal one. Therefore, the problem is how to avoid the obstacle in an efficient way, i.e., to minimize overall time required by the swarm to perform avoidance maneuver, without increasing the velocity significantly, i.e., aggressive acceleration. For example, for an individual agent selecting the nearest edge to bypass the obstacle may be optimum, but for the swarm as a whole it may not be. This is due to the fact that delays occur when the swarm has to deviate from its original trajectory to either avoid an obstacle or go through the available gap between the obstacles and the agents have to slow down, wait, or allow for other neighboring agents to go ahead or merge in the queue as shown in Figure \ref{fig.eg2}. Now it is important to note here that if an obstacle, assuming obstacle is in detection range and both corners are visible, clearly extends towards one side of the swarm does not mean that going for the shortest path will provide optimal results, i.e., minimum time for the last agent to pass through. Here we are calculating the time from when the obstacle is detected till the last agent passes the center of the obstacle, which is our cost function.

To support this claim, we investigate three different scenarios where a swarm faces an obstacle in its way: the first scenario is when the obstacle is inline with the center of the swarm, the second is when the obstacle is to the left of the swarm, and the third is when the obstacle is to the right of the swarm. Figure \ref{fig:responsegraph} shows the timing result of these three scenarios. Here, the agents are divided into two groups, namely: \textbf{Group 1} (N1) deviates from the left side and \textbf{Group 2} (N2) deviates from the right side as shown in Figure \ref{fig:sys}. In this case, the swarm is composed of 8 agents, in a nested V-shaped formation as shown and the results are reported for: 1) \textit{Centered}, when the obstacle's center is inline with the center of the swarm, the optimal result obtained is when \textit{N1} and \textit{N2} both have 4 agents, it took the swarm \textit{28 sec} to pass the obstacle (as shown in Figure \ref{fig:responsegraph}), 2) \textit{Left Shifted}, when the obstacle's center is shifted to left side w.r.t. the swarm's center, the optimal result was acquired with 6 agents in \textit{N1} and 2 agents in \textit{N2}, and 3) \textit{Right Shifted}, when the obstacle's center is shifted to right, the optimal timing for bypassing the obstacle is obtained, i.e., $t_{min}$ = \textit{32sec} with 2 agents in \textit{N1} and 6 agents in \textit{N2}.

It is important to note here that even though it might be possible to reduce the delays by accelerating aggressively to minimize the time delay, however it has an adverse affect on the power consumption of the agent as the minimum power requirement changes \cite{stolaroff_energy_2018}. Therefore, in the performed simulations, the agents maintain a velocity of $v_i$ whenever possible and are allowed to accelerate/decelerate linearly to $v_i\pm \delta$.
%, and deriving it to match our conditions we get the following power to velocity relation:

% \begin{equation}
%     P = T(v sin \gamma + v_i)
% \end{equation}

% where T is the thrust, $\gamma$ is the pitch angle, and $v_i$ is the induced velocity necessary for a given thrust. We know that thrust can be found by the following relation:

% \begin{equation}
%     T = (m_{overall})g + (1/2)\rho v_a^2 C_D A
% \end{equation}

% Where $m_{overall}$ is the overall mass of the agent, including batteries and payloads, $g$ is the gravitational constant, and the second part of the equation is the drag force, where $\rho$ is the density of air, $v_a^2$ is the air speed, $C_D$ is the drag coefficient, and $A$ is the project area. And implicitly solving for induced velocity we get the following relation:

% \begin{equation}
%     v_i = \frac{v_a^2 C_D A}{\pi nD(\sqrt{(vcos\gamma)^2+(vsin\gamma + v_i)^2}}
% \end{equation}

% Where $n$ and $D$ are the number of propellers and diameter of the propeller.

\section{Proposed Approach}
\vspace{-0.15cm}
\begin{figure}[!ht]
    \centering
    \includegraphics[width=0.35\textwidth]{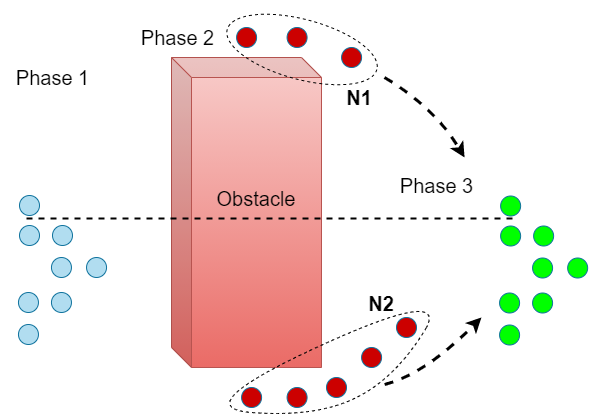}
    \caption{Illustration of 3 phases: Phase 1) the initial phase, Phase 2) system at highest disturbance point, Phase 3) the convergence phase\vspace{-0.2cm} \label{fig:sys}}
\end{figure}
In this section, we describe the proposed Swarm Formation Morphing for Congestion Aware Collision Avoidance algorithm for a swarm of autonomous agents. The overall strategy is to combine the optimal formation morphing, in the presence of obstacles, to avoid overpopulation and the reformation mechanism facilitate the process of efficient navigation of the swarm, Figure \ref{fig:sys}. In the optimal formation morphing for avoidance maneuver, based on the number of obstacles, the population factors are evaluated for the agents of the swarm. Then using the population factor and the time these factors require for avoidance completion, agents are divided into different sets of groups. These set of groups of agents nominate their own respective local leaders of the groups. Once the obstacle avoidance is successfully completed, the swarm at this point remains in the highest disturbed phase. In the convergence phase, thin-plate splines based reformation methodology is presented for bringing the agents back into the intended formation as optimally as possible. Where based on the position vectors, agents are mapped onto the desired formation positions in an optimal manner.

Algorithm \ref{algo1} highlights the global routine of the proposed approach. As an initial assumption, it is considered that the mission is started with already maintained formation shape and the connection is established between the agents already. This top-level algorithm is executed by each agent locally, by utilizing its on-board processing unit. Algorithm \ref{algo1} starts by checking and setting up the leader (global) for the swarm and then proceeds to connecting the follower agents with their respective leaders, if it has not been established yet (Line 2). Then based on the current state, i.e., position, of each agent in swarm, the $Target\_Shape$ of the swarm is initialized (Line 3). Where $Target\_Shape$ represents the future shape of the formation, i.e., the next shape, and tells the next target coordinates each agent should navigate towards. %As a last initialization, the $Detection$ flag is set up to be "$False$", which indicates the detection of any obstacles in the vicinity (Line 8).

\begin{algorithm}
\caption{Global Routine}\label{algo1}
\scriptsize
\begin{algorithmic}[1]
\Procedure{Obstacle Detection \& Navigation()}{}
\State{Leader follower ID assignment;}
\State{Target\_Shape $\gets$ Initialize w.r.t. current state;}
\While{($Detection, d_o, D_{zone}) \gets$ Obstacle Detection() [$Detection$]}
    \State{Grouping process();}
    \State{Leader assignment in local groups;}
    \State{Collision avoidance();}
\EndWhile
\State{Reformation();}
\EndProcedure
\end{algorithmic}
\end{algorithm}\vspace{-0.2cm}
After these initialization, the main loop begins (Lines 4-8), where \textit{Obstacle Detection} is the the first procedure that is executed (Line 4). In case, an obstacle is in the detection range, the procedure locally sets up the global $Detection$ flag. If $Detection == True$, i.e., an obstacle was detected, a certain set of rules are executed (Lines 5-7). First the procedure \textit{Grouping Process()} is called. This procedure is responsible for determining the optimal formation morphing configuration. By utilizing the information provided by \textit{Obstacle Detection()}, i.e., number of obstacles, this procedure calculates the population factors and the respective times they require (approximately) to pass the obstacle, then the most optimal solution is chosen.
%Then if the agent is the leader, based on the characteristics of the obstacle, it approximates escape route trajectories to find the optimal number of agents on either side of the obstacle to avoid over population on either side of the obstacle ultimately reducing the mission time and energy consumption due to the delays which may occur to due agents waiting for other agents to pass (Lines 12-14). Based on the assignments of the agents to both sides (we call left side as \textit{N1} and right side as \textit{N2}), agents select the secondary leader (in \textit{N1} and \textit{N2}) locally by cross-checking the coordinates with each other and the agent closest to the destination declares itself as the leader of the group (Line 15). 
Then based on the calculations and leader determination from the \textit{Grouping Process}, leader assignment is done (line 6). Then the procedure \textit{Collision Avoidance} is called to guide the agents reliably and safely away from the potential collisions (Line 7). And finally, the reformation procedure based on thin-plate spline is called to bring the agents back into the desired formation (Line 9). Its effect is only significant in case the formation has been distorted because of collision avoidance. The importance of point set registration in the reformation process is of significant importance as it is vital to do the mapping between the current and the expected shapes optimally and swiftly.

\subsection{Obstacle Detection}

\begin{figure}[!ht]
    \centering
    \includegraphics[width=0.17\textwidth]{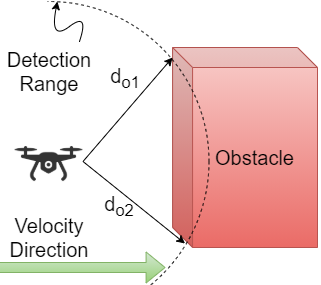}
    \caption{Obstacle Detection\label{fig:detction}}
\end{figure}

The pseudo code for this procedure is specified in Algorithm 2, in which the agent scans for the presence of any objects continuously and the moment an object is detected by the on-board sensor system, the $Detection$ flag is set to \textit{True} (Lines 2-3). Then based on the sensor's feedback, the calculation of the detected object's parameters is done, i.e., the distance at which the object is detected and the angle to it (Line 4), as shown in Figure \ref{fig:detction}.
\begin{algorithm}
\caption{Obstacle Detection}\label{algo2}
\scriptsize
\begin{algorithmic}[1]
\Procedure{Obstacle Detection()}{}
\If{$Obstacle$ in $Detection Range$}
    \State{$Detection$ = True;}
    \State{$d_o$ $\gets$ Calculate the distance and angles to the obstacle;}
    \State{$D\_zone$ $\gets$ calculate the danger zone;}
\EndIf
\EndProcedure
\end{algorithmic}
\end{algorithm}

Considering the velocity at which the agent itself is travelling and the distance to the detected obstacle, the danger zone is defined, beyond which the collision is imminent and can not be avoided (Line 5). The danger zone is defined to adjust the velocities of the agents appropriately based on the braking distance of the agent, as elaborated in the equations below.
We know the distance to the obstacle and the velocity of the agent, then the time to potential impact $t_{imp}$ is calculated by:
\begin{equation}
    t_{imp} = d_{oi} / v
\end{equation}
where $d_{oi}$ is the distance to the object(s) and $v$ is the agent's velocity. And the stopping distance is computed as follows:
\begin{equation}
    d_s = d_r + d_b
\end{equation}
Where $d_s$, $d_r$, and $d_b$ are the stopping distance, reaction distance, and braking distance respectively. Braking distance ($d_b$) and reaction distance ($d_r$) are calculated as follows:
\begin{equation}
    d_b = v^2 / 2 g c_d
\end{equation}
\begin{equation}
    d_r = v t_c
\end{equation}
Where $g$ is the gravitational constant, $c_d$ is the air drag coefficient, and $t_c$ is the time it takes to compute or react.

\subsection{Grouping Process}

In this process, specified in Algorithm \ref{algogrp}, the leader approximates by simulating the time to avoid the obstacle(s) for all the combinations of the agents by utilizing their respective velocities and coordinates. The number of obstacles (\textit{\{obsSet\}}), in the vicinity, is used to calculate the population factor set (\textit{\{popfacSet\}}), Line 2 in Algorithm 3. For instance, if there is one obstacle, then the population factor is two. The available number of population factors can be defined by the following relation:
\begin{equation}
    pf_i = obs_i + 1
\end{equation}
where $pf_i$ is the number of population factor and $obs_i$ is the  number of obstacles in Eq. 5. Then based on the population factor set and the penalty of time, i.e., the group configuration that requires minimum amount of time to pass the obstacle, the swarm is grouped into different set of groups (\textit{\{groupSet\}}) is calculated (Line 3). Afterwards, \textit{\{leaderSet\}} gets the leaders determined from the respective calculated groups, i.e., \textit{\{groupSet\}} (Line 4).

\begin{algorithm}
\caption{Grouping Process}\label{algogrp}
\scriptsize
\begin{algorithmic}[1]
\Procedure{Grouping Process()}{}
\State{$\{popfacSet\}$ $\gets$ Calculate population factor ($\{obsSet\}$);}
\State{$\{groupSet\}$ $\gets$ group ($\{popfacSet\}$);}
\State{$\{leaderSet\} \gets$ determine leader ($\{groupSet\}$);}
%\State{Select group configuration with least TTA;}
\EndProcedure
\end{algorithmic}
\end{algorithm}
\vspace{-0.2cm}

\subsection{Collision Avoidance}

\begin{algorithm}
\caption{Collision Avoidance}\label{algo3}
\scriptsize
\begin{algorithmic}[1]
\Procedure{Collision Avoidance()}{}
\While{$d_o$ $<$ $DetectionRange$}
    \If{obstacle number $>$ 1}
        \State{$gap$ $\gets$ calculate the distance between obstacles;}
        \If{$gap$ $>$ $dist_{safe}$}
            \State{path planning(edges);}\Comment{agent is aligned w.r.t. the gap}
        % \Else 
        %     \State{$plan$ $\gets$ single obstacle;}
        %     \State{path planning(plan);}\Comment{single obstacle scenario}
        \EndIf 
    \Else
        \State{$plan$ $\gets$ single obstacle;}
        \State{path planning(plan);}\Comment{single obstacle scenario}
    \EndIf
\EndWhile
\State{$Detection$ = False;}
\EndProcedure
\end{algorithmic}
\end{algorithm}

The pseudo code, in Algorithm \ref{algo3}, describes the collision avoidance procedure. This procedure is executed when the obstacle is detected and the calculated distance and angles suggest that continuing the trajectory will lead to a collision. It starts by checking if there were multiple obstacles detected (Line 3). In case, the detected obstacles are more than one, the available gap between the obstacles is then calculated (Line 4). If the gap is greater than the defined minimum safe distance (minimum allowed distance on either side of the agent plus agent's dimensions), the agent is aligned to navigate through the obstacles (Lines 5-6). Otherwise, the obstacles are enveloped as one obstacle and path planning done accordingly to bypass a single obstacle.  Or in case, only one obstacle was detected initially, path planning is performed, for a single obstacle, to bypass the obstacle (8-10). For aligning the agent to navigate through the gap between the obstacles and path planning, we utilized and implemented the technique presented in \cite{jy0919}. If the distance to the obstacle is no longer in the detection range, the control is returned to the global routine by resetting the \textit{Detection} flag to \textit{False}.

\subsection{Reformation}

We take inspiration from the technique presented in \cite{jdj2} and base the reformation function by utilizing point set registration \cite{DBLP:journals/corr/abs-0905-2635, 8594514} that is based on a well known technique used to data interpolation and smoothing issues, i.e., thin-plate splines (TPS) \cite{854733}. The amount by which the formation is distorted is assessed by the energy function as shown in Eq. 6.
\begin{equation}
\begin{split}
    E_{TPS}(f) = \sum_{i=1}^{n}||x_i - f(v_i)||^2 + \\ \lambda\iint[(\frac{\partial^2f}{\partial x^2})^2 +2(\frac{\partial^2f}{\partial x\partial y})^2 + (\frac{\partial^2f}{\partial y^2})]dxdy
\end{split}
\label{etps}
\end{equation}
\begin{figure*}[!ht]
\begin{center}
\vspace{0.2cm}
    \subfigure[\label{fig.sim1}]{\includegraphics[width=0.63\columnwidth]{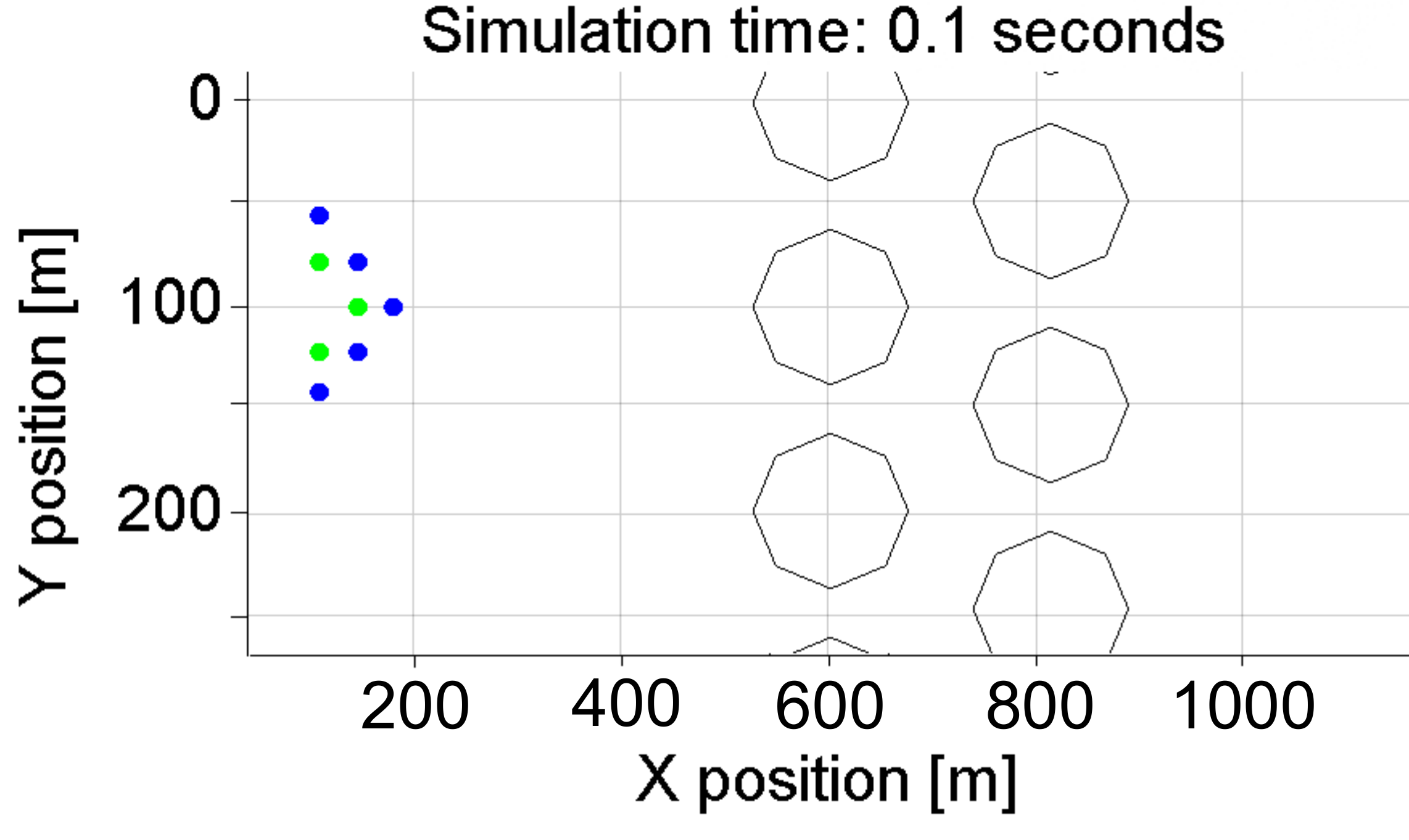}}
    \subfigure[\label{fig.sim2}]{\includegraphics[width=0.63\columnwidth]{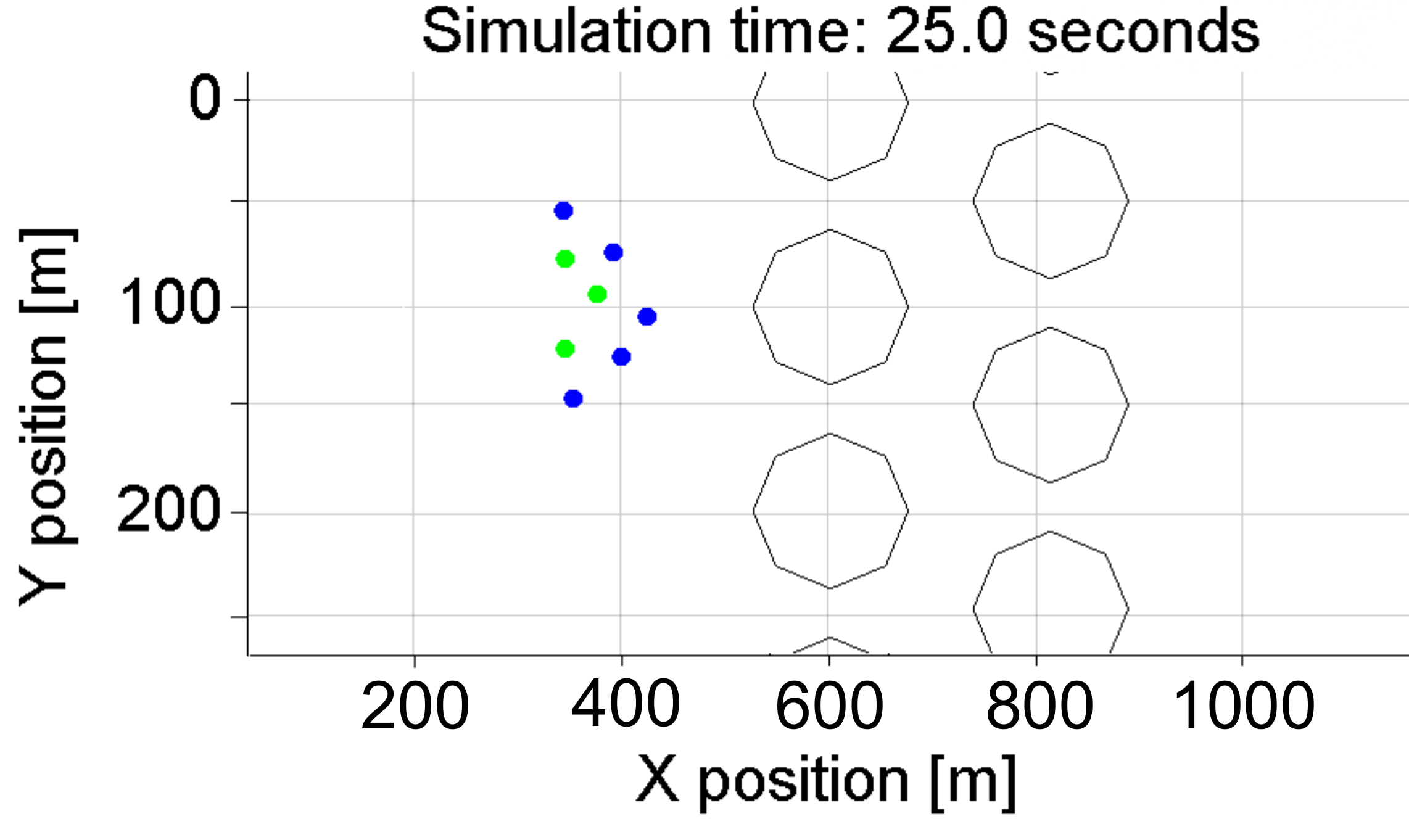}}
    \subfigure[\label{fig.sim3}]{\includegraphics[width=0.63\columnwidth]{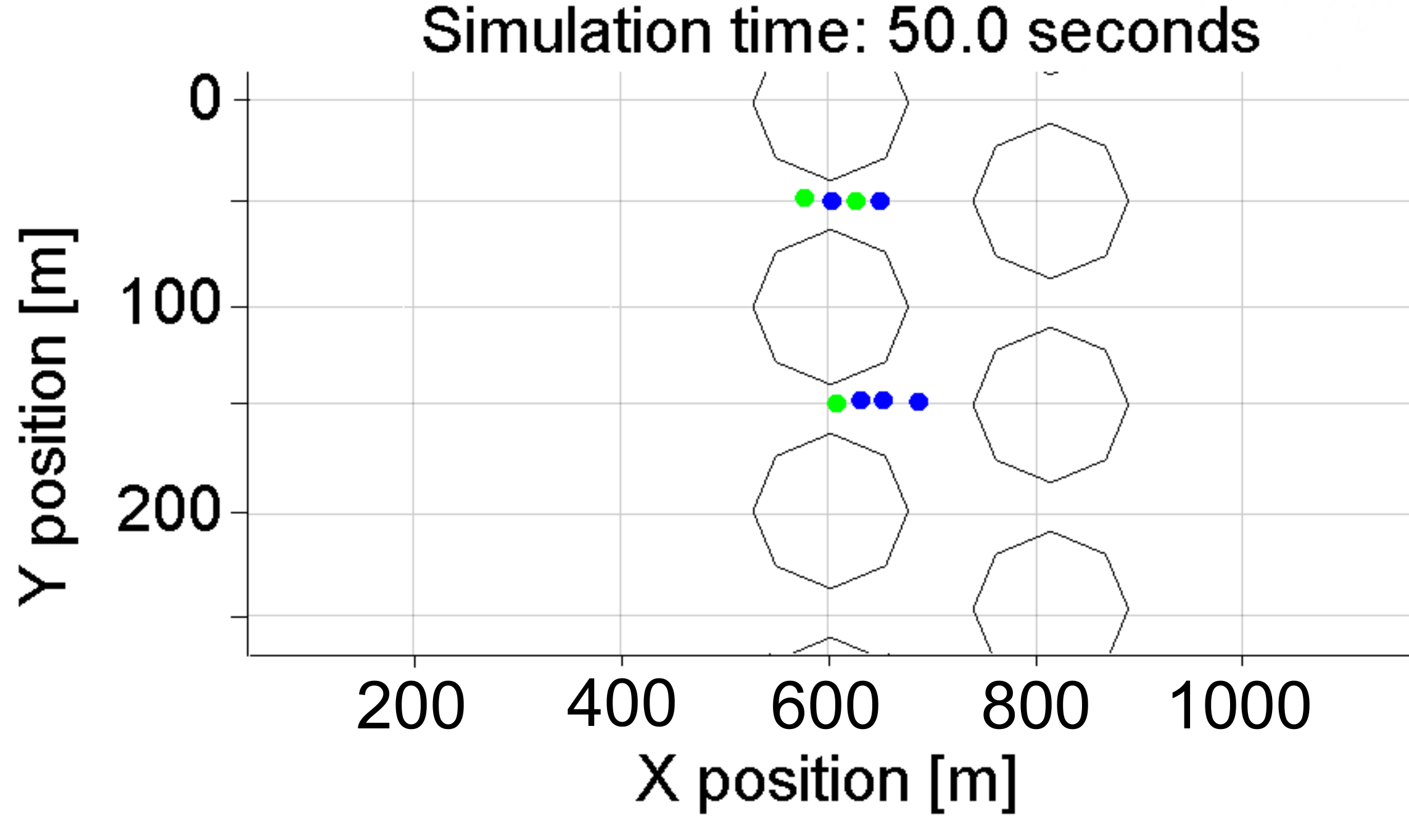}}
    \subfigure[\label{fig.sim4}]{\includegraphics[width=0.63\columnwidth]{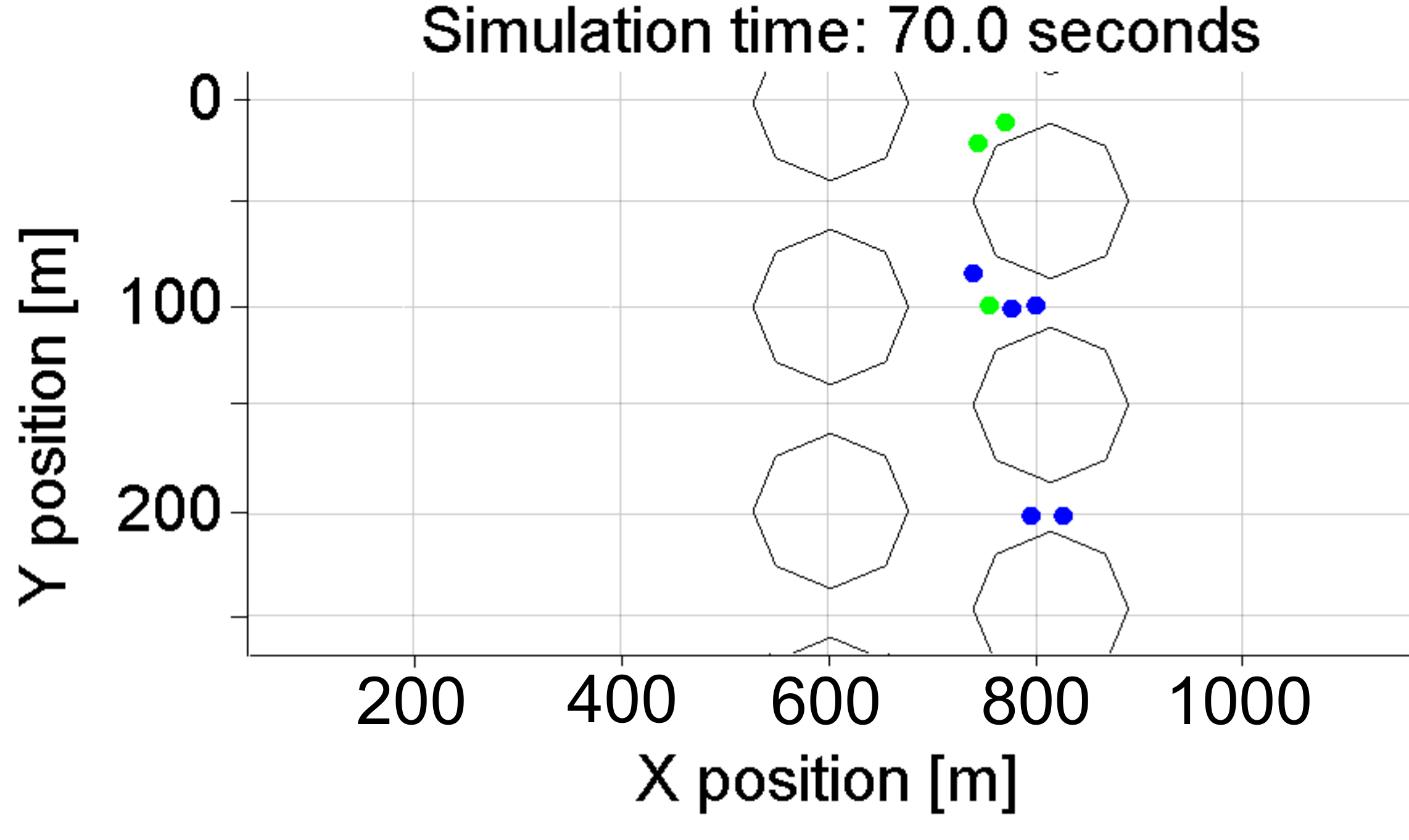}}
    \subfigure[\label{fig.sim5}]{\includegraphics[width=0.63\columnwidth]{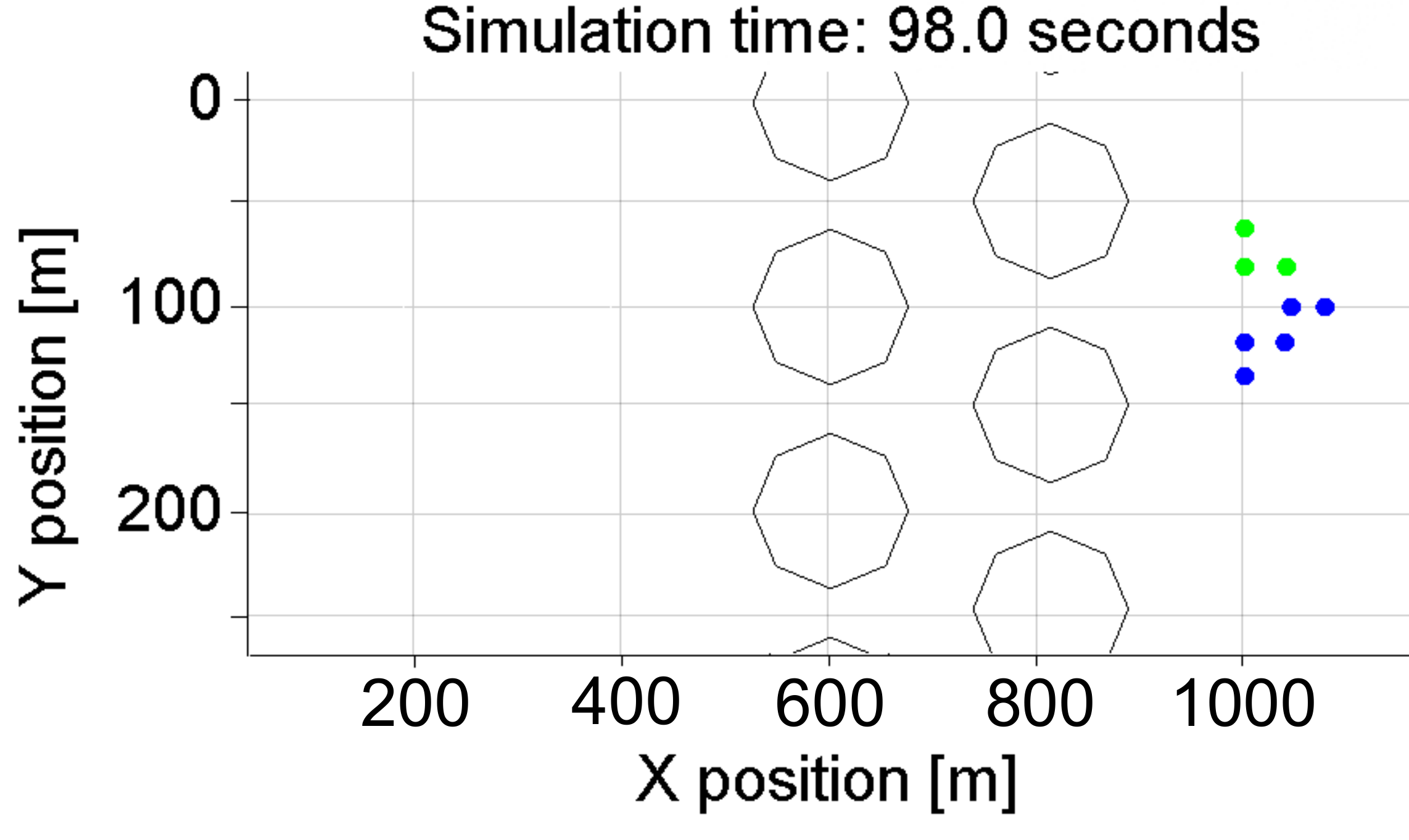}
    }
   \end{center}\vspace{-0.3cm}
  \caption{Simulation snapshots. (a) beginning of the simulation. (b) time = 25s of the simulation. (c) time = 50s of the simulation, swarm navigating through the obstacles. (d) time = 70s, groups divided into further subgroups upon encountering another set of obstacles (e) towards end of simulation.}
  \label{fig:snaps}
\end{figure*}
Where $E_{TPS}$ is the energy function and $\lambda$ is the scaling factor. Mapping a set of points to the corresponding point sets while keeping the the intended \textit{formation} under consideration, is handled by the integral part of the equation. Since, our intention is to only map one set of points over the other and without considering the distorted shape of the swarm, therefore in order to map the closest points we set the scaling factor ($\lambda$) to zero.
\begin{algorithm}
\caption{Reformation}\label{algo4}
\scriptsize
\begin{algorithmic}[2]
\Procedure{Reformation()}{}
\While{Agents have not REACHED new coordinates}
    \State{$Next\_Location$ = Compute the next position of the swarm;}
    \State{$Agent(i)$ = Compute new coordinates for each agent;}
    \State{TPS($Next\_Location$); \Comment{Minimization of temperature func.}}
\EndWhile
\EndProcedure
\end{algorithmic}
\end{algorithm}

The overview of the TPS-based reformation function is provided in Algorithm \ref{algo4}. This procedure starts by computing the next, i.e., the future, location of each agent based on the present coordinates of the agents (Line 3). Then agents are assigned new coordinates based on the determined new location (Line 4). Then, to perform reformation as optimally as possible, these determined values are passed to the temperature minimization function based on TPS, for bringing the agents to their respective updated locations as optimally as possible (Line 5). As soon as every agent has reached its respective new location or coordinates, the control is returned to the global routine.

\section{SIMULATION \& RESULTS}

Assumptions and initial conditions considered in this work are defined as follows:

\begin{enumerate}
    \item all agents are at the same altitude
    \item agents accelerate or decelerate linearly
    \item the position vectors of the agents are obtained by utilizing on-board localization techniques
    \item the communication channel is ideal, i.e., without information loss and delays
    \item the computational or reaction time of the agent is considered to be negligible, and the reaction distance ($d_r$) is zero.
\end{enumerate}

For visualizing agents in the simulation, point mass particle model is used, therefore the point mass particle's equations of motion are utilized in this work. SwarmLab: a MATLAB Drone Swarm Simulator \cite{soria2020} is used for visualization purposes (Figure \ref{fig:snaps}). 

\begin{figure}[!ht]
    \centering
    \vspace*{0.2cm}
    \includegraphics[width=0.45\textwidth]{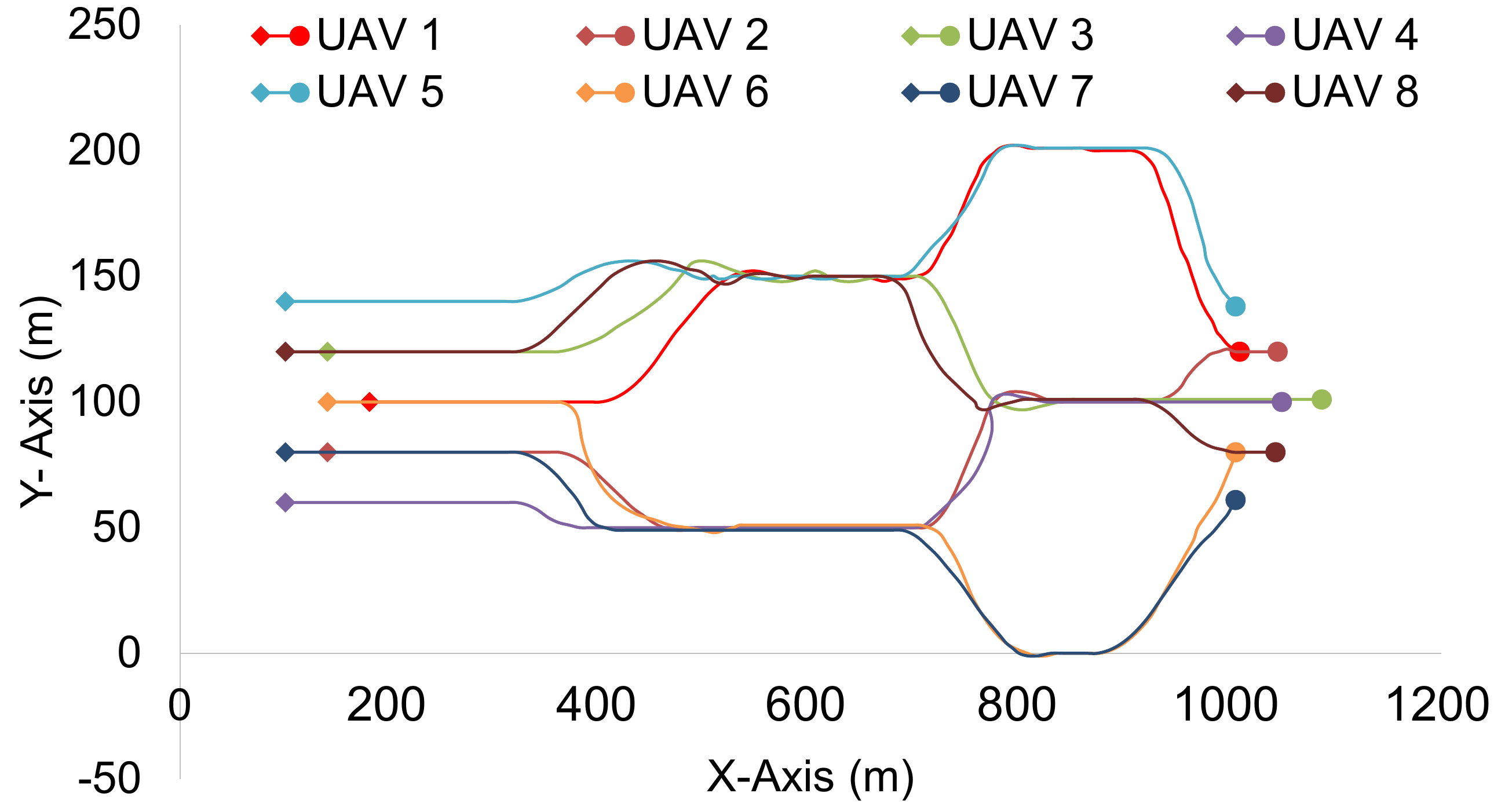}
    \caption{Trace of Overall Movement of the Agents. Here starting locations of all the agents are represented by the diamond shape ($\diamond$) and the final locations of the agents are represented by a dot shape ($\bullet$) \label{fig:rsult}}
\end{figure}

The mission starts, Figure \ref{fig.sim1}, with agents already in a defined nested V-shaped formation moving towards the destination in an open environment. Figure \ref{fig.sim2} shows the disturbed formation at \textit{Simulation time = 25s} when the obstacle has been detected and the agents have started deviating to avoid the potential collision. In Figure \ref{fig.sim3}, the formation disturbance is at maximum, the agents, grouped to avoid congestion on either side (\textit{N1}, \textit{N2})\footnote{Group labelling is done for illustration and explanation purposes. Depending on the scenario, agents can be divided into several group and subgroup sets \{groupSet\}}, while keeping the minimum safe distance from the obstacle and from each other, are bypassing the obstacle. As soon as the agents pass the obstacle, the second set of obstacles is detected by both groups locally. As shown in Figure \ref{fig.sim4}, the agents divide themselves locally into further sub-groups (we call \textit{N11}, \textit{N12} and \textit{N21}, \textit{N22} for sub groups from \textit{N1} and \textit{N2} respectively) to avoid congesting either side. The agents in \textit{N12} and agents in \textit{N21} are navigating towards same route to bypass the obstacles. In this case, the agent farthest ahead takes precedence and in the similar manner they merge to form a queue formation to navigate through space between the obstacles. Figure \ref{fig.sim5}, shows the swarm's reformation once there is no obstacle in the detection range of the agents. The trace for the overall movement of the agents throughout the mission is shown in Figure \ref{fig:rsult}, where morphing of the formation is visible, through a forest like environment. In the figure, the starting points/positions of the agents are denoted by a "diamond" shape and the final positions are denoted by a solid circle.

\begin{figure}[!ht]
    \centering
    \includegraphics[width=0.50\textwidth]{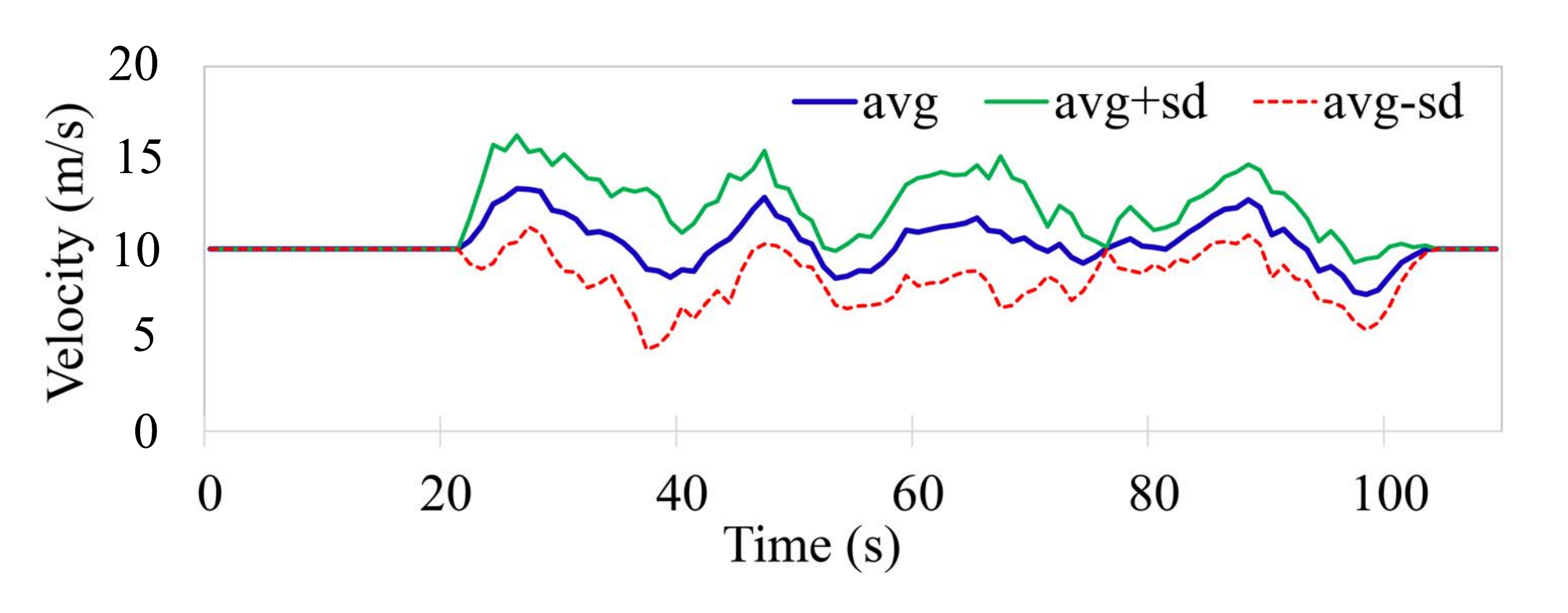}\vspace{-0.4cm}
    \caption{Report of the average Velocity, average $\pm$ standard deviation of the swarm \label{fig:aveV}}\vspace{-0.3cm}
\end{figure}

\begin{figure}[!ht]
    \centering
    \includegraphics[width=0.50\textwidth]{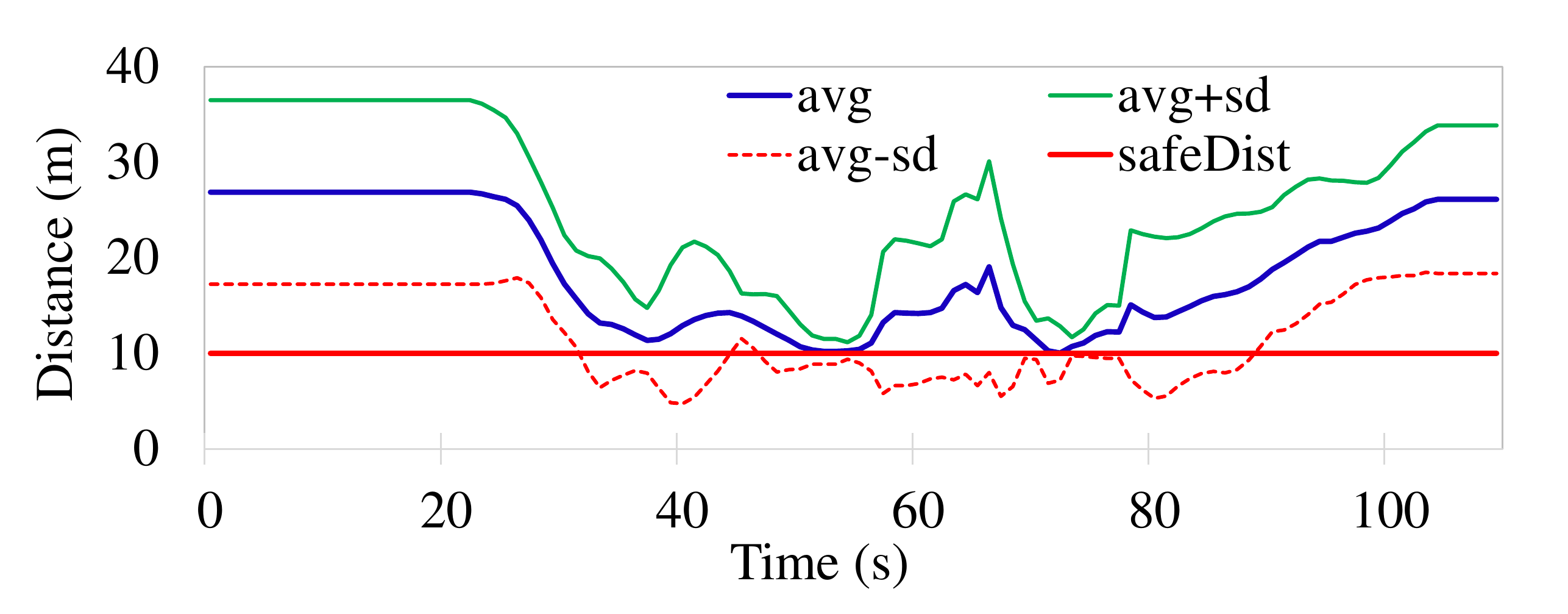}\vspace{-0.4cm}
    \caption{Average distance maintained by the agents from mission start to obstacle avoidance to mission end, average $\pm$ standard deviation of the swarm  \label{fig:aveD}}
\end{figure}

Figure \ref{fig:aveV} shows the average velocity of the swarm as a whole, Figure \ref{fig:aveD} shows the average distance maintained by the agents in the swarm; standard deviation of the velocity and distance is also plotted for reference. The non-aggressive variance in the average velocity is due to the fact that agents have to slow down to provide enough space for another agent, deviate to bypass the obstacle, slow down if the agent in front is slowing down (Figure \ref{fig.sim3}), and maintain tight queue formation while going through obstacles.

\vspace{-0.3cm}
\vspace{-0.1cm}
In Figure \ref{fig:aveD}, the solid red line shows the defined minimum safe distance in the performed experiment, which is the minimum distance agents maintain with each other in the disturbance phase, i.e., while performing collision avoidance. The peak at around \textit{t = 43s}, is due to the change in immediate respective leaders due to the \textit{grouping process}, Figure \ref{fig.sim3}. The second sudden change in the average distance maintained, at \textit{t = 66s}, is due to the encounter with second set of obstacles and the previous groups getting divided into further groups and resultant change in immediate leaders through \textit{grouping process}. From around \textit{t = 80s} onwards, the agents are coming back into the desired formation shape and maintaining the formation defined inter-agent distance. There is a negligible error of 0.8m in the average distance maintained between the agents when reformation is completed.

In  order  to  estimate  the  energy  saving  effect  of  the proposed  approach,  we  consider  energy  consumption  of  the swarm of eight drones while bypassing a single obstacle, as discussed in section II and depicted in Figure  \ref{fig:egfig}. In the following discussion, we use the results of \cite{8663615}, where total power required by a drone weighing 20 Newton, having four blades with rotor radius of 40 cm is plotted against drone’s flying speed. We have selected the nominal speed as 10m/s, which is close approximation of the Maximum Endurance speed calculated by \cite{8663615}. To further enhance endurance, we perform gradual acceleration and deceleration with maximum speed fixed at 20 m/s, the power consumption is seen to rise drastically at speeds above this value, Figure 2 of \cite{8663615} refers. The total energy consumed to perform a maneuver can be found by integrating the instantaneous powers over the whole flight time, from start time ($t_s$) to finish time ($t_f$). Since our algorithm works in discrete time steps, we calculate each drone's the total energy consumption as shown in Eq. 7, and sum up individual results to yield total energy consumption of the swarm as a whole.
\begin{equation}
    E_{total} = \sum_{t=t_s}^{t_f}P(t) \Delta t
\end{equation}
Employing the proposed approach, the total energy consumed by the swarm was 54.111 kJ, whereas utilizing the shortest path algorithm the swarm consumed 62.084 kJ, resulting in 14.7\% higher energy consumption.

\section{CONCLUSIONS}

In this paper, we present a methodology for finding an optimal solution to (1) avoid congestion that may happen in a swarm of autonomous agents while avoiding collisions with obstacles and to (2) bring the agents back into the desired formation shape after evading an obstacle. In the proposed method, an agent population control factor is considered in relation to the obstacle(s) in the vicinity, with a time constraint for the disturbance phase. In this approach, the leader of the swarm takes a centralized decision by utilizing the parameters of the agents (coordinates, velocities) and obstacles to find the population factor, and based on that the grouping configuration of the swarm is determined. Then it selects the group setup which provides the shortest overall congestion delay. Afterwards, in the convergence phase of the proposed methodology, a thin-plate splines based technique is utilized to optimally bring the agents back into the intended formation by mapping the closest agents to the nearest points of the desired formation. We demonstrated via simulations that by utilizing the congestion control approach we can minimize the delays, save time, and consequently minimize the overall energy consumption of the swarm.

In our future work, we aim to refine the proposed methodology by considering non-negligible computation times and communication delays in calculating realistic reaction distances. %Another direction for future work is to explore distributed decision making and switch between distributed and centralized decision making depending on the surrounding environment.

\bibliographystyle{IEEEtran}

\bibliography{ref}

\begin{thebibliography}{10}
\providecommand{\url}[1]{#1}
\csname url@samestyle\endcsname
\providecommand{\newblock}{\relax}
\providecommand{\bibinfo}[2]{#2}
\providecommand{\BIBentrySTDinterwordspacing}{\spaceskip=0pt\relax}
\providecommand{\BIBentryALTinterwordstretchfactor}{4}
\providecommand{\BIBentryALTinterwordspacing}{\spaceskip=\fontdimen2\font plus
\BIBentryALTinterwordstretchfactor\fontdimen3\font minus
  \fontdimen4\font\relax}
\providecommand{\BIBforeignlanguage}[2]{{%
\expandafter\ifx\csname l@#1\endcsname\relax
\typeout{** WARNING: IEEEtran.bst: No hyphenation pattern has been}%
\typeout{** loaded for the language `#1'. Using the pattern for}%
\typeout{** the default language instead.}%
\else
\language=\csname l@#1\endcsname
\fi
#2}}
\providecommand{\BIBdecl}{\relax}
\BIBdecl

\bibitem{Hamann2018}
\BIBentryALTinterwordspacing
H.~Hamann, \emph{Introduction to Swarm Robotics}.\hskip 1em plus 0.5em minus
  0.4em\relax Cham: Springer International Publishing, 2018, pp. 1--32.
  [Online]. Available: \url{https://doi.org/10.1007/978-3-319-74528-2\_1}
\BIBentrySTDinterwordspacing

\bibitem{dorigo2004swarm}
M.~Dorigo and A.~F. Roosevelt, ``Swarm robotics,'' in \emph{Special Issue”,
  Autonomous Robots}.\hskip 1em plus 0.5em minus 0.4em\relax Citeseer, 2004.

\bibitem{7989678}
A.~{Tagliabue}, M.~{Kamel}, S.~{Verling}, R.~{Siegwart}, and J.~{Nieto},
  ``Collaborative transportation using mavs via passive force control,'' in
  \emph{Proc. IEEE International Conference on Robotics and Automation (ICRA)},
  2017, pp. 5766--5773.

\bibitem{8682048}
H.~{Shakhatreh}, A.~H. {Sawalmeh}, A.~{Al-Fuqaha}, Z.~{Dou}, E.~{Almaita},
  I.~{Khalil}, N.~S. {Othman}, A.~{Khreishah}, and M.~{Guizani}, ``Unmanned
  aerial vehicles (uavs): A survey on civil applications and key research
  challenges,'' \emph{IEEE Access}, vol.~7, pp. 48\,572--48\,634, 2019.

\bibitem{1678135}
B.~{Grocholsky}, J.~{Keller}, V.~{Kumar}, and G.~{Pappas}, ``Cooperative air
  and ground surveillance,'' \emph{IEEE Robotics Automation Magazine}, vol.~13,
  no.~3, pp. 16--25, 2006.

\bibitem{6385551}
J.~{Alonso-Mora}, M.~{Schoch}, A.~{Breitenmoser}, R.~{Siegwart}, and
  P.~{Beardsley}, ``Object and animation display with multiple aerial
  vehicles,'' in \emph{Proc. IEEE/RSJ International Conference on Intelligent
  Robots and Systems}, 2012, pp. 1078--1083.

\bibitem{8764393}
S.~A.~S. {Mohamed}, M.~{Haghbayan}, T.~{Westerlund}, J.~{Heikkonen},
  H.~{Tenhunen}, and J.~{Plosila}, ``A survey on odometry for autonomous
  navigation systems,'' \emph{IEEE Access}, vol.~7, pp. 97\,466--97\,486, 2019.

\bibitem{9197184}
H.~{Nguyen}, T.~{Dang}, and K.~{Alexis}, ``The reconfigurable aerial robotic
  chain: Modeling and control,'' in \emph{Proc. IEEE International Conference
  on Robotics and Automation (ICRA)}, 2020, pp. 5328--5334.

\bibitem{jdj2}
J.~N. {Yasin}, S.~A.~S. {Mohamed}, M.~H. {Haghbayan}, J.~{Heikkonen},
  H.~{Tenhunen}, M.~M. {Yasin}, and J.~{Plosila}, ``Energy-efficient formation
  morphing for collision avoidance in a swarm of drones,'' \emph{IEEE Access},
  vol.~8, pp. 170\,681--170\,695, 2020.

\bibitem{jdj1}
J.~N. {Yasin}, S.~A.~S. {Mohamed}, M.~{Haghbayan}, J.~{Heikkonen},
  H.~{Tenhunen}, and J.~{Plosila}, ``Unmanned aerial vehicles (uavs): Collision
  avoidance systems and approaches,'' \emph{IEEE Access}, vol.~8, pp.
  105\,139--105\,155, 2020.

\bibitem{7850263}
S.~{Mostaghim}, C.~{Steup}, and F.~{Witt}, ``Energy aware particle swarm
  optimization as search mechanism for aerial micro-robots,'' in \emph{2016
  IEEE Symposium Series on Computational Intelligence (SSCI)}, 2016, pp. 1--7.

\bibitem{Majd_2020}
A.~Majd, M.~Loni, G.~Sahebi, and M.~Daneshtalab, ``Improving motion safety and
  efficiency of intelligent autonomous swarm of drones,'' \emph{Drones},
  vol.~4, no.~3, p.~48, Aug 2020.

\bibitem{8711119}
K.~{Narayanan}, V.~{Honkote}, D.~{Ghosh}, and S.~{Baldev}, ``Energy efficient
  communication with lossless data encoding for swarm robot coordination,'' in
  \emph{Proc. 32nd International Conference on VLSI Design and 2019 18th
  International Conference on Embedded Systems (VLSID)}, 2019, pp. 525--526.

\bibitem{10.1007/978-3-030-49778-1_28}
J.~N. Yasin, S.~A.~S. Mohamed, M.-H. Haghbayan, J.~Heikkonen, H.~Tenhunen, and
  J.~Plosila, ``Navigation of autonomous swarm of drones using translational
  coordinates,'' in \emph{Advances in Practical Applications of Agents,
  Multi-Agent Systems, and Trustworthiness. The PAAMS Collection}, Y.~Demazeau,
  T.~Holvoet, J.~M. Corchado, and S.~Costantini, Eds.\hskip 1em plus 0.5em
  minus 0.4em\relax Cham: Springer International Publishing, 2020, pp.
  353--362.

\bibitem{tseng2017autonomous}
\BIBentryALTinterwordspacing
C.-M. Tseng, C.-K. Chau, K.~Elbassioni, and M.~Khonji, ``Autonomous recharging
  and flight mission planning for battery-operated autonomous drones,'' 2017.
  [Online]. Available: \url{https://arxiv.org/abs/1703.10049}
\BIBentrySTDinterwordspacing

\bibitem{8284558}
B.~N. {Chand}, P.~{Mahalakshmi}, and V.~P.~S. {Naidu}, ``Sense and avoid
  technology in unmanned aerial vehicles: A review,'' in \emph{Proc.
  International Conference on Electrical, Electronics, Communication, Computer,
  and Optimization Techniques (ICEECCOT)}, 2017, pp. 512--517.

\bibitem{stolaroff_energy_2018}
\BIBentryALTinterwordspacing
J.~K. Stolaroff, C.~Samaras, E.~R. O’Neill, A.~Lubers, A.~S. Mitchell, and
  D.~Ceperley, ``\BIBforeignlanguage{en}{Energy use and life cycle greenhouse
  gas emissions of drones for commercial package delivery},''
  \emph{\BIBforeignlanguage{en}{Nature Communications}}, vol.~9, no.~1, p. 409,
  Feb. 2018, number: 1 Publisher: Nature Publishing Group. [Online]. Available:
  \url{https://www.nature.com/articles/s41467-017-02411-5}
\BIBentrySTDinterwordspacing

\bibitem{jy0919}
J.~N. Yasin, M.-H. Haghbayan, J.~Heikkonen, H.~Tenhunen, and J.~Plosila,
  ``Formation maintenance and collision avoidance in a swarm of drones,'' in
  \emph{Proc. International Symposium on Computer Science and Intelligent
  Control}, ser. ISCSIC 2019.\hskip 1em plus 0.5em minus 0.4em\relax New York,
  NY, USA: Association for Computing Machinery, 2019.

\bibitem{DBLP:journals/corr/abs-0905-2635}
\BIBentryALTinterwordspacing
A.~Myronenko and X.~B. Song, ``Point-set registration: Coherent point drift,''
  \emph{CoRR}, vol. abs/0905.2635, 2009. [Online]. Available:
  \url{http://arxiv.org/abs/0905.2635}
\BIBentrySTDinterwordspacing

\bibitem{8594514}
P.~{Guo}, W.~{Hu}, H.~{Ren}, and Y.~{Zhang}, ``Pcaot: A manhattan point cloud
  registration method towards large rotation and small overlap,'' in
  \emph{Proc. IEEE/RSJ International Conference on Intelligent Robots and
  Systems (IROS)}, Oct 2018, pp. 7912--7917.

\bibitem{854733}
{Haili Chui} and A.~{Rangarajan}, ``A new algorithm for non-rigid point
  matching,'' in \emph{Proceedings IEEE Conference on Computer Vision and
  Pattern Recognition. CVPR 2000 (Cat. No.PR00662)}, vol.~2, June 2000, pp.
  44--51 vol.2.

\bibitem{soria2020}
\BIBentryALTinterwordspacing
E.~{Soria}, F.~{Schiano}, and D.~{Floreano}, ``Swarmlab: a matlab drone swarm
  simulator,'' in \emph{Proc. IEEE/RSJ International Conference on Intelligent
  Robots and Systems (IROS)}, 2020. [Online]. Available:
  \url{https://arxiv.org/abs/2005.02769}
\BIBentrySTDinterwordspacing

\bibitem{8663615}
Y.~{Zeng}, J.~{Xu}, and R.~{Zhang}, ``Energy minimization for wireless
  communication with rotary-wing uav,'' \emph{IEEE Transactions on Wireless
  Communications}, vol.~18, no.~4, pp. 2329--2345, 2019.

\end{thebibliography}

\end{document}